\documentclass[runningheads]{llncs}
\usepackage{marvosym}
\usepackage{amsmath,amssymb,amsfonts}
\usepackage{algorithm,algorithmic}
\usepackage{booktabs} 
\usepackage{multirow}
\usepackage{array}
\usepackage[T1]{fontenc}
\usepackage{graphicx,verbatim}
\usepackage{hyperref}
 
\begin{document}

\title{Fusion-E2Pulse: A Multimodal Event-RGB Fusion
Network for Non-contact Pulse Wave
Reconstruction}
\titlerunning{Fusion-E2Pulse}

\author{Qian Feng\inst{1} \and
Hao Guo\inst{1} \and
Yan Niu\inst{1} \and
Zhenhuan Xu\inst{2}\textsuperscript{\Letter} \and
Yidi Li\inst{1}\textsuperscript{\Letter}}
\index{Feng, Qian}
\index{Guo, Hao}
\index{Niu, Yan}
\index{Xu, Zhenhuan}
\index{Li, Yidi}

\authorrunning{Q. Feng et al.}

\institute{College of Computer Science and Technology,\\ Taiyuan University of Technology, China \and
College of Artificial Intelligence,\\
Taiyuan University of Technology, China\\
\email{\{xuzhenhuan, liyidi\}@tyut.edu.cn}}

  
\maketitle 
\begin{abstract}
Non-contact pulse wave reconstruction hinges on the precise recovery of waveform morphology, including the dicrotic notch. Conventional Red-Green-Blue (RGB)-based methods, which extract physiological signals from recorded facial videos, are constrained by the integral imaging mechanism of standard cameras, where the exposure process induces a smoothing effect that attenuates subtle vascular pulsation details. Conversely, neuromorphic event cameras, while offering exceptional sensitivity to intensity fluctuations, are inherently susceptible to noise and artifacts induced by minor motion. To exploit the synergy between frame-based integration and event-based differential sensing, we propose a novel multimodal network named Fusion-E2Pulse. This framework utilizes filtered RGB signals as structural priors to suppress motion artifacts, while leveraging the high sensitivity of event streams to recover fine-grained morphological details. Experimental results demonstrate that Fusion-E2Pulse achieves state-of-the-art performance, effectively balancing noise suppression and morphological fidelity, achieving a mean absolute error of 0.78 bpm for heart rate estimation, a waveform correlation of 0.89, and a systolic phase duration error of 16.74 ms, validating its efficacy in reconstructing fine-grained pathological features.

\end{abstract}

\section{Introduction}
\label{sec:introduction}
Cardiovascular disease remains a leading cause of global mortality, requiring effective screening and long-term monitoring to mitigate its impact \cite{song2021pulsegan,wang2025non}. As a vital physiological window, pulse wave morphology reflects the dynamic interplay between the heart and the peripheral vascular system. Through pulse wave analysis, clinicians can obtain fundamental indicators such as Heart Rate (HR) and Heart Rate Variability (HRV). More importantly, the precise extraction of morphological features—notably the dicrotic notch—provides critical pathological information regarding vascular resistance and arterial stiffness, which are key predictors of early-stage cardiovascular deterioration \cite{adami2024rppg,wang2025non}. While traditional contact-based sensors offer high precision, they are often impractical for sensitive scenarios such as neonatal care or burn patient monitoring \cite{oronti12025optimizing,figl2008registration,zhao2024pulsenet}. Consequently, remote photoplethysmography (rPPG) has emerged as a promising non-contact alternative due to its low hardware cost and accessibility \cite{mcduff2023camera,wang2024camera}. This field has evolved significantly, ranging from classical signal processing methods such as CHROM~\cite{de2013robust}, ICA~\cite{poh2010non}, and POS~\cite{wang2015novel} to state-of-the-art (SOTA) deep learning architectures like PhysNet~\cite{yu2019remote} and RhythmFormer~\cite{zouRhythmFormerExtractingPatterned2025}.

Despite these advancements, standard Red-Green-Blue (RGB)-based methods face an inherent physical bottleneck. These methods rely on capturing Photoplethysmography (PPG) signals—a measure of blood volume changes via skin color variations—from time-series pixel values in facial video sequences. The integral imaging mechanism of standard cameras acts as a physical low-pass filter by accumulating photons over fixed exposure windows, which inevitably smooths out the subtle, rapid pulsations essential for capturing subtle morphological structures \cite{de2013robust}. The intrinsic advantages of neuromorphic event cameras precisely compensate for these deficiencies. Characterized by acute sensitivity to intensity fluctuations and high dynamic range, event cameras asynchronously encode pixel-level intensity changes to acutely capture the subtle variations caused by skin micro-vibrations \cite{gallego2020event}. Recent studies have confirmed the feasibility of non-contact HR monitoring using event cameras \cite{nakamura2025novel,jagtap2023heart}, highlighting their potential to record subtle physiological dynamics. However, the differential sensing of event cameras \cite{lv2024structural} is a double-edged sword; while providing high sensitivity, it is highly susceptible to non-physiological clutter triggered by minute motions, compromising HR estimation robustness.

In this study, we propose an innovative multimodal network named \textbf{Fusion-E2Pulse} to achieve robust noise suppression and high-fidelity waveform reconstruction. By establishing a complete framework that synergizes RGB structural stability with event dynamic sensitivity, we bridge the gap between global stability and local precision. 
The principal contributions are summarized as follows: (1) we propose a novel fusion paradigm that integrates RGB structural priors with event streams to effectively suppress motion-induced artifacts; (2) we establish a time-frequency adversarial supervision framework with synergistic optimization objectives, integrating spectral and morphological constraints to ensure high-fidelity physiological recovery; and (3) we perform extensive validation on the Event-based Multimodal Physiological Dataset (EMPD) \cite{feng_2026_18765701}, where Fusion-E2Pulse establishes new SOTA performance in HR estimation and enables precise reconstruction of subtle physiological landmarks.

\section{Method}
\label{sec:method}
\begin{figure}[!t]
    \centering
    \includegraphics[width=\linewidth]{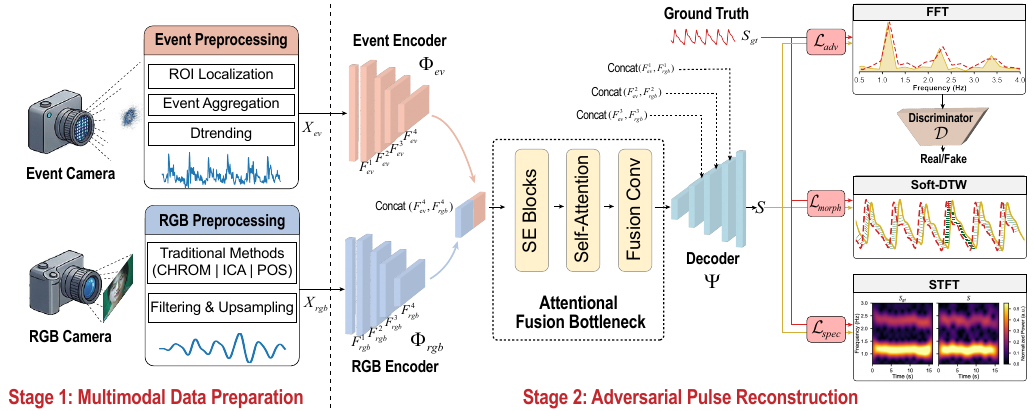} 
    \caption{The architecture of the proposed Fusion-E2Pulse framework.}
    \label{fig:architecture}
\end{figure}

To implement the aforementioned synergy between RGB structural stability and event dynamic sensitivity, the Fusion-E2Pulse framework is organized into a two-stage pipeline consisting of multimodal data preparation and adversarial pulse reconstruction, as illustrated in Fig. \ref{fig:architecture}. We formulate pulse wave reconstruction as a multimodal sequence-to-sequence generation task, where the objective is to recover the underlying PPG signal ($S_{gt}$)—serving as the ground truth recorded via a fingertip pulse oximeter—from heterogeneous sensor data. By bridging the gap between global stability and local precision, the framework transforms raw sensor captures into high-fidelity physiological waveforms.

\subsection{Multimodal Data Preparation}

The raw input modalities comprise asynchronous event streams captured from the radial artery of the wrist by an event camera and recorded RGB video sequences. To handle modal heterogeneity and prepare the data for the generative network, we apply distinct preprocessing pipelines to derive the final processed signals $X_{ev} \in \mathbb{R}^{1 \times T}$ and rPPG signals $X_{rgb} \in \mathbb{R}^{1 \times T}$, where $T$ denotes the time window length. Unlike traditional contact-based sensors, our framework extracts physiological information from the spatio-temporal pixel variations within video sequences and sparse event spikes, leveraging their unique sensing mechanisms to ensure robust reconstruction.

\begin{figure}[!t]
    \centering
    \includegraphics[width=\linewidth]{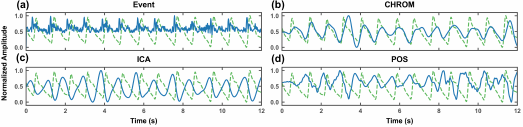}
    \caption{Visualization of synchronized multimodal inputs. Inputs are overlaid with the ground-truth PPG signal (green dashed). (a) Event signal. (b)--(d) RGB signals extracted via CHROM, ICA, and POS.}
    \label{fig:inputs}
\end{figure}

For the event stream, we first apply Region of Interest (ROI) localization followed by temporal aggregation to convert asynchronous events into a synchronous continuous signal matching the sampling frequency of the ground-truth PPG ($60\text{Hz}$), with detrending applied to remove motion-induced drifts. Simultaneously, for the RGB stream, rPPG waveforms are extracted from video via traditional rPPG algorithms (CHROM \cite{de2013robust}, ICA \cite{poh2010non}, and POS \cite{wang2015novel}), which leverage the specific absorption characteristics of oxygenated hemoglobin in the visible spectrum. These signals are band-pass filtered and linearly interpolated for temporal alignment. As visualized in Fig. \ref{fig:inputs}, the resulting processed signals $X_{ev}$ and $X_{rgb}$ exhibit significant morphological complementarity: $X_{ev}$ displays rich fine-grained dynamics that acutely capture subtle vascular pulsations, whereas $X_{rgb}$ maintains a robust low-frequency periodic structure but is overly smoothed due to the integral imaging effect. Finally, both signals are Z-score standardized \cite{liu2023rppg,abdi2007z} to facilitate stable fusion.

\subsection{Adversarial Pulse Reconstruction}
\textbf{Dual-Stream Encoder.} To extract complementary features, the encoder employs a dual-stream architecture. The event branch ($\Phi_{ev}$) extracts fine-grained local features $F_{ev}^i$ via cascaded 1D convolutions, where $i$ is the layer index. Complementarily, the RGB branch ($\Phi_{rgb}$) extracts low-frequency periodic structural features $F_{rgb}^i$. Both branches output semantic features $F_{ev}^{L}$ and $F_{rgb}^{L}$ at the deep bottleneck ($L$ being the total layers) to avoid early fusion interference.

\noindent \textbf{Attentional Fusion Bottleneck.} To address modal heterogeneity, we introduce Squeeze-and-Excitation (SE) \cite{hu2018squeeze} and Self-Attention (SA) \cite{munchmeyer2021transformer} mechanisms. Features are first concatenated as $F_{cat} = \text{Concat}(F_{ev}^{L}, F_{rgb}^{L})$. An SE module then dynamically calibrates channel weights to produce $F_{se}$, allowing adaptive weighting of event and RGB features based on signal quality. Subsequently, SA captures global temporal dependencies to yield $F_{sa}$. Finally, a fusion convolution (1$\times$1 Conv) reduces the channel dimensionality to produce the final fused feature $F_{fused}$ for the decoder.

\noindent \textbf{Decoder.} The decoder ($\Psi$) upsamples the fused feature $F_{fused}$ via transposed convolutions to restore temporal resolution. 
Crucially, skip connections link the shallow features ($F_{ev}^i, F_{rgb}^i$) from the encoders to the corresponding decoder layers. 
This fuses texture details with semantic information, finally generating the reconstructed pulse waveform $S = \Psi(F_{fused})$.

\noindent \textbf{Discriminator.} Unlike standard GANs that operate directly on time-domain signals, our discriminator $\mathcal{D}$ transforms the waveform into the frequency domain via Fast Fourier Transform (FFT).
It analyzes the magnitude spectrum of the signal, which allows the network to explicitly distinguish the fundamental HR frequency and harmonic structures from noise.
This spectral discrimination strategy guides the generator to synthesize realistic physiological periodicity without over-constraining the temporal phase.

\noindent \textbf{Optimization Objectives.} To ensure consistency in both time-domain topology and frequency-domain distribution, we train our network with a composite loss:
$\mathcal{L}_{total} = \mathcal{L}_{adv} + \lambda_{morph}\mathcal{L}_{morph} + \lambda_{spec}\mathcal{L}_{spec}$. 
We employ an adversarial loss ($\mathcal{L}_{adv}$) based on Wasserstein GAN \cite{gulrajani2017improved} with gradient penalty to constrain the distribution of the generated pulse waves. To ensure morphological fidelity and robustness against phase shifts, we utilize the Soft-DTW loss ($\mathcal{L}_{morph}$) \cite{cuturi2017soft}, which provides a measure of temporal alignment. Finally, a time-frequency spectral loss ($\mathcal{L}_{spec}$) based on Short-Time Fourier Transform (STFT) \cite{allen1997short} is applied to suppress noise and enforce consistency in the spectrogram domain.

\begin{figure}[!b]
    \centering
    \includegraphics[width=\linewidth]{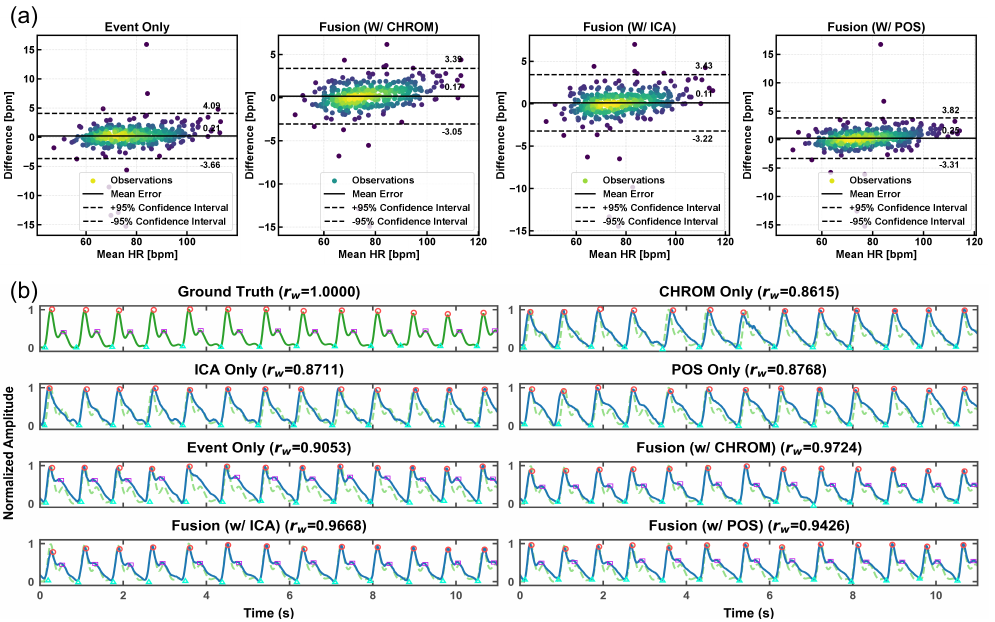}
    \caption{\textbf{(a)} Bland-Altman plots showing the agreement between the estimated and ground-truth HR. The plots display the difference (y-axis) against the mean (x-axis), with dashed lines indicating the 95\% Limits of Agreement (LoA); \textbf{(b)} Waveform morphology analysis.}
    \label{fig:bland_altman}
\end{figure}

\section{Experiment and Results}
\label{sec:experiment}
\subsection{Datasets and Implementation Details}
To comprehensively evaluate Fusion-E2Pulse, we utilize the EMPD \cite{feng_2026_18765701} dataset, comprising 193 multimodal recordings (68s each) from 83 subjects, including synchronized facial RGB, wrist-pulse events, and ground-truth PPG. We adopt two evaluation protocols: (1) time split, partitioning each 68s record into training (48s), validation (8s), and testing (12s) segments to assess temporal continuity; and (2) record split, randomly partitioning the 193 records into training (70\%), validation (12\%), and testing (18\%) sets to evaluate generalization across samples. Experiments are implemented in PyTorch on an NVIDIA RTX 4060 GPU using the Adam optimizer ($\beta_1=0.5, \beta_2=0.999$) with an initial learning rate of $2 \times 10^{-4}$. Loss weights are set to $\lambda_{morph}=1.0$ and $\lambda_{spec}=10.0$ to balance constraints. 
Evaluation metrics for heart rate estimation accuracy include Mean Absolute Error (MAE, in \textit{bpm}), Root Mean Square Error (RMSE, in \textit{bpm}), Mean Absolute Percentage Error (MAPE, in \%), and the Pearson coefficient ($r_h$). For morphological fidelity, we evaluate the waveform Pearson coefficient ($r_w$), Signal-to-Noise Ratio (SNR, in \textit{dB}), and the Systolic/Diastolic Phase Duration (SPD/DPD) and Pulse Width (PWD) measured in \textit{ms} \cite{tong2023detail,korpas2009parameters}.

\begin{table}[!t]
  \centering
  \caption{Quantitative comparison of the proposed Fusion-E2Pulse against its single-modal counterparts.}
  \label{tab:results_combined}
  \footnotesize
  \begin{tabular}{lccccccccc}
    \toprule
    \multirow{2}{*}{\textbf{Method}} & \multicolumn{4}{c}{\textbf{HR}} & \multicolumn{5}{c}{\textbf{Morphology}} \\
    \cmidrule(lr){2-5} \cmidrule(lr){6-10}
     & \textbf{MAE} & \textbf{RMSE} & \textbf{MAPE} & \textbf{$r_h$} & \textbf{SPD} & \textbf{DPD} & \textbf{PWD} & \textbf{SNR} & \textbf{$r_w$} \\
    \midrule
    \multicolumn{10}{c}{\textit{\textbf{Time Split}}} \\
    \midrule
    Event & 0.87 & 2.68 & 1.18 & 0.977 & 22.50 & 29.22 & 17.93 & 11.46 & 0.883 \\
    CHROM & 5.10 & 9.50 & 6.16 & 0.713 & 28.49 & 60.04 & 64.50 & 6.41 & 0.543 \\
    ICA   & 6.63 & 11.36 & 8.31 & 0.543 & 22.67 & 69.47 & 73.41 & 6.15 & 0.498 \\
    POS   & 5.74 & 10.27 & 6.98 & 0.656 & 26.26 & 58.84 & 65.64 & 6.40 & 0.498 \\
    w/ CHROM & 0.82 & \textbf{2.51} & 1.14 & \textbf{0.980} & \textbf{16.07} & \textbf{19.24} & \textbf{15.98} & \textbf{11.93} & 0.894 \\
    w/ ICA & 0.89 & 2.65 & 1.26 & 0.978 & 18.52 & 21.85 & 18.82 & 11.81 & \textbf{0.895} \\
    \textbf{w/ POS} & \textbf{0.78} & \textbf{2.51} & \textbf{1.05} & \textbf{0.980} & 16.74 & 21.85 & 17.93 & 11.89 & 0.891 \\
    \midrule
    \multicolumn{10}{c}{\textit{\textbf{Record Split}}} \\
    \midrule
    Event & 1.00 & 3.84 & 1.27 & 0.940 & 16.71 & 20.08 & 15.43 & 12.64 & 0.904 \\
    CHROM & 4.50 & 8.43 & 6.00 & 0.692 & 21.88 & 55.26 & 56.41 & 6.93 & 0.558 \\
    ICA   & 5.76 & 10.59 & 7.42 & 0.517 & 21.61 & 63.38 & 65.22 & 6.77 & 0.538 \\
    POS   & 5.11 & 9.51 & 6.96 & 0.612 & 20.51 & 58.47 & 59.51 & 6.86 & 0.550 \\
    w/ CHROM & 0.94 & 3.65 & 1.22 & 0.945 & 16.64 & 19.46 & 14.97 & \textbf{12.93} & 0.909 \\
    w/ ICA & 1.00 & 3.80 & 1.25 & 0.941 & 17.48 & 20.77 & 17.19 & 12.79 & 0.907 \\
    \textbf{w/ POS} & \textbf{0.83} & \textbf{2.88} & \textbf{1.10} & \textbf{0.966} & \textbf{16.59} & \textbf{18.05} & \textbf{14.32} & 12.79 & \textbf{0.915} \\
    \bottomrule
  \end{tabular}
\end{table}

\subsection{Comparative Analysis with Single-Modal Counterparts}
We benchmark Fusion-E2Pulse against various configurations to validate the efficacy of multimodal integration. These counterparts include: (1) event-only, a single-modal network variant (denoted as Event in our experimental tables) utilizing only wrist-pulse event streams as input without RGB guidance; and (2) RGB-only: pulse signals extracted solely from facial video sequences via classic rPPG algorithms, including CHROM \cite{de2013robust}, ICA \cite{poh2010non}, and POS \cite{wang2015novel}.
The proposed fusion variants, denoted as w/ CHROM, w/ ICA, and w/ POS, represent the full Fusion-E2Pulse architecture integrating event micro-dynamics with the corresponding structural priors derived from these RGB-based methods. This setup allows for a direct assessment of how different structural references influence the reconstruction of morphological details.

\begin{table}[!b]
  \centering
  \caption{Ablation study of Fusion-E2Pulse loss functions and architecture components. (Top) Impact of different optimization objectives under the time split. (Bottom) Robustness of fusion strategies and component contributions under the record split.}
  \label{tab:ablation_combined}
  \footnotesize 
  \begin{tabular}{llccccccccc}
    \toprule
    \multirow{2}{*}{\textbf{Setting}} & \multirow{2}{*}{\textbf{Method}} & \multicolumn{4}{c}{\textbf{HR}} & \multicolumn{5}{c}{\textbf{Morphology}} \\
    \cmidrule(lr){3-6} \cmidrule(lr){7-11}
     & & \textbf{MAE} & \textbf{RMSE} & \textbf{MAPE} & \textbf{$r_h$} & \textbf{SPD} & \textbf{DPD} & \textbf{PWD} & \textbf{SNR} & \textbf{$r_w$} \\
    \midrule
    \multicolumn{11}{c}{\textit{\textbf{Time Split}}} \\
    \midrule
    \multirow{4}{*}{$\mathcal{L}_{adv}$} & Event & 21.35 & 37.26 & 24.91 & 0.294 & -- & -- & -- & -- & -- \\
     & w/ CHROM & 29.27 & 43.97 & 35.95 & 0.154 & -- & -- & -- & -- & -- \\
     & w/ ICA & 26.14 & 40.37 & 32.37 & 0.072 & -- & -- & -- & -- & -- \\
     & w/ POS & 17.71 & 23.34 & 20.95 & 0.071 & -- & -- & -- & -- & -- \\
    \midrule
    \multirow{4}{*}{\shortstack{$\mathcal{L}_{spec}$\\+$\mathcal{L}_{adv}$}} & Event & 1.02 & 4.02 & 1.32 & 0.949 & 16.50 & 20.32 & 16.70 & 11.18 & 0.881 \\
     & w/ CHROM & 0.97 & 2.70 & 1.31 & 0.979 & 18.90 & 25.52 & 21.36 & 11.56 & 0.883 \\
     & w/ ICA & 1.09 & 3.61 & 1.50 & 0.958 & 19.31 & 23.51 & 17.90 & 11.31 & 0.884 \\
     & w/ POS & 0.97 & 2.84 & 1.29 & 0.974 & 17.99 & 21.54 & 20.41 & 11.31 & 0.880 \\
    \midrule
    \multirow{4}{*}{\shortstack{$\mathcal{L}_{morph}$\\+$\mathcal{L}_{adv}$}} & Event & 1.11 & 3.10 & 1.49 & 0.970 & 20.92 & 28.97 & 19.99 & 11.41 & 0.880 \\
     & w/ CHROM & 1.04 & 2.84 & 1.37 & 0.974 & 18.63 & 25.48 & 19.55 & 11.24 & 0.875 \\
     & w/ ICA & 1.22 & 3.63 & 1.59 & 0.958 & 18.53 & 24.03 & 19.52 & 11.28 & 0.876 \\
     & w/ POS & 1.07 & 3.02 & 1.41 & 0.970 & 18.50 & 24.13 & 20.36 & 11.77 & 0.882 \\
    \midrule
    \multirow{4}{*}{\textbf{Full}} & Event & 0.87 & 2.68 & 1.18 & 0.977 & 22.50 & 29.22 & 17.93 & 11.46 & 0.883 \\
     & w/ CHROM & 0.82 & \textbf{2.51} & 1.14 & \textbf{0.980} & \textbf{16.07} & \textbf{19.24} & \textbf{15.98} & \textbf{11.93} & 0.894 \\
     & w/ ICA & 0.89 & 2.65 & 1.26 & 0.978 & 18.52 & 21.85 & 18.82 & 11.81 & \textbf{0.895} \\
     & \textbf{w/ POS} & \textbf{0.78} & \textbf{2.51} & \textbf{1.05} & \textbf{0.980} & 16.74 & 21.85 & 17.93 & 11.89 & 0.891 \\
    \midrule
    \multicolumn{11}{c}{\textit{\textbf{Record Split}}} \\
    \midrule
    \multirow{4}{*}{$\mathcal{L}_{adv}$} & Event & 20.87 & 31.86 & 25.85 & 0.158 & -- & -- & -- & -- & -- \\
     & w/ CHROM & 28.33 & 42.61 & 36.62 & 0.078 & -- & -- & -- & -- & -- \\
     & w/ ICA & 18.57 & 28.92 & 24.07 & -0.049 & -- & -- & -- & -- & -- \\
     & w/ POS & 28.23 & 42.58 & 36.66 & 0.066 & -- & -- & -- & -- & -- \\
    \midrule
    \multirow{4}{*}{\shortstack{$\mathcal{L}_{spec}$\\+$\mathcal{L}_{adv}$}} & Event & 0.98 & 3.69 & 1.31 & 0.944 & 28.01 & 32.19 & 14.92 & 10.02 & 0.722 \\
     & w/ CHROM & 0.91 & 3.57 & 1.17 & 0.948 & 19.09 & 24.41 & \textbf{13.85} & 8.31 & 0.820 \\
     & w/ ICA & 0.97 & 3.66 & 1.25 & 0.945 & 45.72 & 51.18 & 20.17 & 7.56 & 0.645 \\
     & w/ POS & 1.08 & 4.43 & 1.47 & 0.919 & 18.42 & 23.44 & 14.68 & 7.78 & 0.836 \\
    \midrule
    \multirow{4}{*}{\shortstack{$\mathcal{L}_{morph}$\\+$\mathcal{L}_{adv}$}} & Event & 1.11 & 3.10 & 1.49 & 0.970 & 20.92 & 28.97 & 19.99 & 11.41 & 0.880 \\
     & w/ CHROM & 1.04 & 2.84 & 1.37 & 0.974 & 18.63 & 25.48 & 19.55 & 11.24 & 0.875 \\
     & w/ ICA & 1.22 & 3.63 & 1.59 & 0.958 & 18.53 & 24.03 & 19.52 & 11.28 & 0.876 \\
     & w/ POS & 1.07 & 3.02 & 1.41 & 0.970 & 18.50 & 24.13 & 20.36 & 11.77 & 0.882 \\
    \midrule
    \multirow{4}{*}{\textbf{Full}} 
     & Event & 1.00 & 3.84 & 1.27 & 0.940 & 16.71 & 20.08 & 15.43 & 12.64 & 0.904 \\
     & w/ CHROM & 0.94 & 3.65 & 1.22 & 0.945 & 16.64 & 19.46 & 14.97 & \textbf{12.93} & 0.909 \\
     & w/ ICA & 1.00 & 3.80 & 1.25 & 0.941 & 17.48 & 20.77 & 17.19 & 12.79 & 0.907 \\
     & \textbf{w/ POS} & \textbf{0.83} & \textbf{2.88} & \textbf{1.10} & \textbf{0.966} & \textbf{16.59} & \textbf{18.05} & 14.32 & 12.79 & \textbf{0.915} \\
    \bottomrule
  \end{tabular}
\end{table}

\noindent \textbf{Performance on Time Split.} To evaluate the fundamental reconstruction capability, we benchmark the models using the time split (Table~\ref{tab:results_combined}). Fusion-E2Pulse (w/ POS) achieves the lowest MAE (0.78 bpm) while maintaining high waveform correlation ($r_w \approx 0.89$), significantly outperforming single-modal counterparts. Qualitatively, our method reconstructs sharp dicrotic notches (Fig.~\ref{fig:bland_altman}(b), $r_w \approx 0.97$) and demonstrates the tightest Limits of Agreement (LoA) in Bland-Altman analysis (Fig.~\ref{fig:bland_altman}(a)). Critically, the errors remain consistently low across the varying heart rate range shown on the x-axis, indicating robustness against physiological variations.

\noindent \textbf{Generalization on Record Split.} To assess robustness against unseen subjects, we evaluate the models using the record split (Table~\ref{tab:results_combined}). All single-modal counterparts experience performance degradation; notably, the event-only MAE deteriorates from 0.87 bpm to 1.00 bpm, indicating sensitivity to individual-specific motion patterns and skin textures. However, Fusion-E2Pulse maintains superior stability. The fusion variant (w/ POS) achieves an MAE of 0.83 bpm and an SPD error of 16.59 ms. This suggests that the RGB structural prior acts as a universal structural constraint, guiding the event branch to adapt to diverse physiological characteristics and ensuring high-fidelity reconstruction even for unseen individuals.

\subsection{Ablation Study}
\noindent \textbf{Impact of Optimization Objectives.} We analyze the contribution of each loss component under the time split (Table~\ref{tab:ablation_combined}). The configuration trained solely with $\mathcal{L}_{adv}$ fails to converge (MAE $=$ 17.71 bpm), highlighting the difficulty of stabilizing GANs on unconstrained physiological signals. The introduction of $\mathcal{L}_{spec}$ proves critical, drastically reducing MAE to 1.00 bpm by anchoring the fundamental frequency. The final addition of $\mathcal{L}_{morph}$ via Soft-DTW further refines the waveform topology, optimizing SNR and morphological metrics (SPD/DPD) to their best values.

\noindent \textbf{Robustness of Fusion Strategy.} We further investigate the fusion effectiveness under the record split (Table~\ref{tab:ablation_combined}). Consistent with time-split results, the inclusion of structural priors (CHROM, ICA, POS) consistently improves performance over the event-only configuration. Specifically, the fusion variant (w/ POS) improves the SNR from 12.64 dB (event-only) to 12.79 dB. Notably, even without the full loss constraints, the structural prior provides essential regularization, preventing the event branch from overfitting to subject-specific noise. This confirms that the attentional fusion bottleneck successfully extracts complementary features—RGB for global stability and event for local dynamic precision.

\section{Conclusion}
\label{sec:conclusion}
In this work, we propose Fusion-E2Pulse, a novel multimodal framework synergizing neuromorphic event cameras and RGB video for high-fidelity pulse reconstruction. By integrating RGB structural priors with event micro-dynamics under a time-frequency adversarial supervision framework, our approach overcomes single-modal limitations such as exposure smoothing and noise. Comprehensive evaluations on the EMPD dataset demonstrate superior performance in HR estimation and the recovery of fine-grained morphological biomarkers. Future research will address Pulse Transit Time (PTT) \cite{wang2024camera} alignment challenges inherent in cross-site sensing and optimize the framework for real-time edge deployment through unsupervised domain adaptation and model compression.


\bibliographystyle{splncs04}
\bibliography{refer}
\end{document}